\documentclass[10pt,twocolumn,letterpaper]{article}

\usepackage[pagenumbers]{cvpr} 

\usepackage[accsupp]{axessibility}
\usepackage{graphicx}
\usepackage{amsmath}
\usepackage{amssymb}
\usepackage{booktabs}
\usepackage{mathtools}
\usepackage{multirow}
\usepackage{colortbl}
\usepackage{soul}
\DeclarePairedDelimiter\abs{\lvert}{\rvert}%
\usepackage[pagebackref,breaklinks,colorlinks]{hyperref}

\usepackage[capitalize]{cleveref}
\crefname{section}{Sec.}{Secs.}
\Crefname{section}{Section}{Sections}
\Crefname{table}{Table}{Tables}
\crefname{table}{Tab.}{Tabs.}

\def\cvprPaperID{40} 
\def\confName{CVPR}
\def\confYear{2023}

\begin{document}

\title{CLVOS23: A Long Video Object Segmentation Dataset for Continual Learning}
\author{Amir Nazemi,
Zeyad Moustafa,
Paul Fieguth \\
 University of Waterloo, Waterloo, Ontario, Canada\\
{\tt\small  {\{amir.nazemi,zeyad.moustafa,paul.fieguth\}}@uwaterloo.ca}
}
\maketitle

\begin{abstract}
Continual learning in real-world scenarios is a major challenge. A general continual learning model should have a constant memory size and no predefined task boundaries, as is the case in semi-supervised Video Object Segmentation (VOS), where continual learning challenges particularly present themselves in working on long video sequences. In this article, we first formulate the problem of semi-supervised VOS, specifically online VOS, as a continual learning problem, and then secondly provide a public VOS dataset, CLVOS23, focusing on continual learning. Finally, we propose and implement a regularization-based continual learning approach on LWL, an existing online VOS baseline, to demonstrate the efficacy of continual learning when applied to online VOS and to establish a CLVOS23 baseline. We apply the proposed baseline to the Long Videos dataset as well as to two short video VOS datasets, DAVIS16 and DAVIS17. To the best of our knowledge, this is the first time that VOS has been defined and addressed as a continual learning problem. The proposed CLVOS23 dataset has been released at~\url{https://github.com/Amir4g/CLVOS23}.
\end{abstract}

\section{Introduction}
\label{sec:intro}

The goal of Video Object Segmentation (VOS) is to accurately extract a target object at the pixel level from each frame of a given video.
In general, there are two categories of VOS solutions: semi-supervised or one-shot VOS, in which the ground-truth masks of the target objects are given in at least one frame at inference time, and unsupervised VOS, in which the VOS model knows nothing about the objects.

Among semi-supervised VOS approaches, online VOS approaches~\cite{robinson2020learning,mao2021joint,bhat2020learning} update a part of the VOS model based on the evaluated frames and estimated masks. The idea is that videos contain relevant information beyond just the given frame's mask, which a model can exploit by learning during the evaluation process.

Online model learning, {\em while} a video is being analyzed, leads to questions regarding how effectively the model learns from frame to frame, particularly when some aspect of the video looks different than what had been given in the ground-truth frame.  This leads to the domain of  continual learning, which is a type of machine learning where a model is trained on a sequence of tasks, and is expected to continuously improve its performance on each new task while retaining its ability to perform well on previously-learned tasks.

The current state-of-the-art semi-supervised and specifically online VOS methods~\cite{robinson2020learning,mao2021joint,bhat2020learning} perform well on VOS datasets with {\em short} videos (up to a few seconds or 100 frames in length) such as DAVIS16~\cite{perazzi2016benchmark}, DAVIS17~\cite{perazzi2016benchmark}, and YouTube-VOS18~\cite{xu2018youtube}.  However, most of these methods do not retain their expected performance on long videos, such as those in the Long Videos dataset~\cite{liang2020video} as shown in the XMem paper~\cite{cheng2022xmem}. The question of the poor performance of online VOS on long videos has not been investigated in the VOS field, nor addressed through continual learning.

Continual learning methods are typically tested on classification datasets, like MNIST~\cite{lecun1998gradient}, CIFAR10~\cite{krizhevsky2009learning}, and Imagenet~\cite{5206848}, or on datasets specifically designed for continual learning, such as Core50~\cite{lomonaco2017core50}. 
The classification dataset is fed to the model as a sequential stream of data in online continual learning methods~\cite{aljundi2019gradient}.
In contrast to the aforementioned datasets and test scenarios, long video object segmentation has numerous real-world applications, such as video summarization, human-computer interaction, and autonomous vehicles~\cite{yao2020video}.

In this paper, we formulate and address the inefficient performance of the online VOS approaches on long videos as an online continual learning problem.
Moreover, we propose a new long-video object segmentation dataset for continual learning (CLVOS23), as a much more realistic and significantly greater challenge for testing VOS methods on long videos.  As a baseline, we propose a Regularization-based (prior-focused) Continual Learning (RCL) solution to improve online VOS.

\section{Related work}
\label{sec:related_work}
Semi-supervised VOS methods try to maximize the benefit from whatever information is given, normally the first frame of the video. Early solutions in the literature \cite{caelles2017one,perazzi2017learning} fine-tuned a pretrained VOS on the given information in a video at evaluation time. In contrast, current state-of-the-art solutions attempt to benefit from previously evaluated frames and make use of an allocated memory to preserve that information from preceding frames in segmenting the current frame.
The so called memory-based VOS approaches~\cite{zhou2019enhanced,oh2019video,robinson2020learning,joint,bhat2020learning,cheng2022xmem} also are categorised into two streams, matching-based and online: 
\begin{itemize}
    \item Matching-based VOS methods~\cite{hu2021learning,xie2021efficient,cheng2021rethinking,yang2021associating,lin2021query,liu2022global,seong2021hierarchical} match the representations of previous frames, stored in memory, with the corresponding features extracted from the current frame. 
    \item Online VOS~\cite{caelles2017one,VoigtlaenderL17,robinson2020learning,bhat2020learning,liu2020meta} update (fine-tune) a small model based on the features and estimated masks of preceding frames.
\end{itemize}
Continual learning~\cite{aljundi2019continual,hsu2018re,yang2021dystab} is a sequential learning process where the data sequence may come from different domains and tasks; thus, a model is learning from data where distribution drift~\cite{gama2014survey} may occur suddenly or gradually.
Catastrophic forgetting is the key challenge in continual learning and it was first defined on neural networks~\cite{mccloskey1989catastrophic,ratcliff1990connectionist} when a neural network model is trained on a sequence of tasks, but has access to the training data for only the current task.
In such circumstances, the model learning process is inclined to frequently update those parameters which are heavily influenced by data from the current task, leading to previously-learned tasks to be partially forgotten. The concept of catastrophic forgetting was also defined on other machine learning models~\cite{erdem2005ensemble}. There are three different approaches to catastrophic forgetting: prior-focused (regularization-based)  \cite{de2021continual,chen2021overcoming}, likelihood-focused (rehearsal-based)  \cite{atkinson2021pseudo,castro2018end,wu2019large,zhao2020maintaining}, and hybrid (ensemble) approaches~\cite{lee2016dual,rusu2016progressive}. 

Elastic Weight Consolidation (EWC)~\cite{kirkpatrick2017overcoming} and Memory Aware Synopses (MAS)~\cite{aljundi2018memory} are two examples of prior-focused methods that employ regularization during training to limit the change of previously learned weights. These methods assume that previously learned task weights can serve as a prior for the current network weights, which are in charge of learning new tasks.
Through the use of a penalty term in the loss function, these methods aim to preserve the significant parameters from preceding tasks.

Likelihood-focused (rehearsal) techniques concentrate on minimizing the model's loss function by taking into account historical information.
Examples include deep generative replay (DGR)~\cite{shin2017continual} and variational generative replay (VGR)~\cite{farquhar2018towards}, which keep previous data or train generative models on earlier tasks prior to training the new task.
Generative Adversarial Networks (GANs) are used in~\cite{shin2017continual} to produce data from each task as samples to be used during the training of a new task.

Finally, as their name implies, hybrid methods seek to combine the benefits of prior-focused and likelihood-focused techniques.
As an example, Variational Continual Learning (VCL)~\cite{nguyen2017variational} combines the posterior from the previous task (i.e., the prior to the current task) with information about the new task (i.e., its likelihood). 

The solution proposed  in this article is a Regularization-based Continual Learning (RCL) approach, drawing its motivation from EWC~\cite{kirkpatrick2017overcoming}.

\section{Problem formulation}
\label{sec:problem_formulation}

\begin{figure*}[t]
	\begin{center}
		\includegraphics[scale = 0.22]{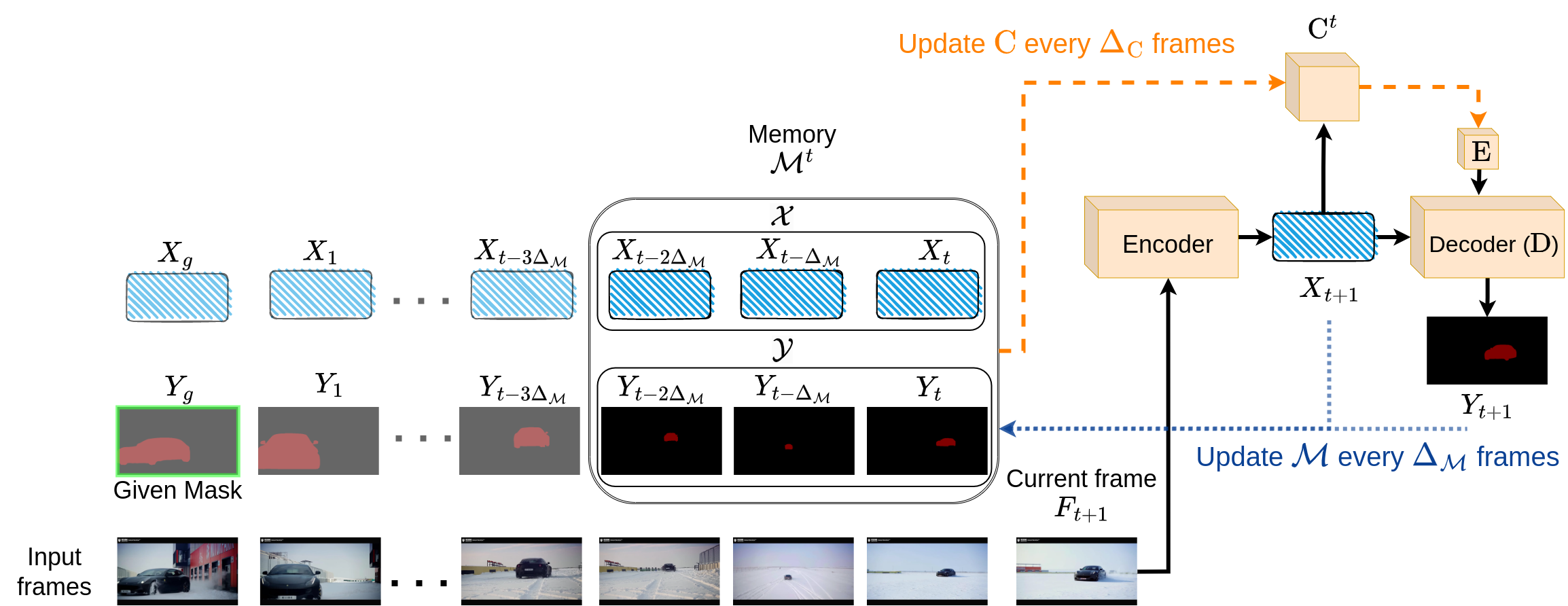}
	\end{center}
	\caption{General online VOS framework: The target model $\mathrm{C^{t-\Delta_{\mathrm{C}}}}$ is updated on memory \(\mathcal{M}
 ^t\) to form $\mathrm{C^{t}}$. The target model $\mathrm{C}$ is initialized based on the given ground truth mask \(Y_g\) and its associated feature \(X_g\). The memory \(\mathcal{M}^{t}\) is updated every \(\Delta_{\mathrm{M}}\) time steps (video frames) with new information $(X_{t+1},Y_{t+1})$. The dashed lines show how the target model $\mathrm{C}$ is updated based on memory \(\mathcal{M}\) every \(\Delta_{\mathrm{C}}\) frames, and the dotted lines show memory update. Our proposed methods focus on the target model component ($\mathrm{C}$) of the framework. The frame images used in the figure are taken from the ``car'' video in the proposed CLVOS23 dataset.}
	\label{fig1}
\end{figure*}
An online VOS model $O_\Xi$~\cite{robinson2020learning,mao2021joint,bhat2020learning} is first trained offline to minimize the following loss function and to learn the model parameters \(\Xi\):
\begin{align}
\label{eq:main-VOS-loss}
\Xi = \mathop{\arg \min}_{\Xi'}\mathcal{L}(O_{\Xi'}(F),Y).
\end{align}
In ~\cref{eq:main-VOS-loss}, $\mathcal{L}$ is usually a pixel-wise cross entropy loss~\cite{chen2014semantic}, \(F\) is an image frame and $Y$ is the segmented mask in which each pixel of $F$ is labeled, based on the number of objects in the video sequence. For example, in the case of single-object video, $Y$ is just a binary foreground/background mask.
An online VOS model typically has a U-Net encoder-decoder structure~\cite{ronneberger2015u}, and further comprises the following pieces:  
\begin{enumerate}
    \item A pretrained encoder, extracting feature $X$ from each frame $F$;
    \item A memory $\mathcal{M} = \{\mathcal{X},\mathcal{Y}\}$, storing features \(\mathcal{X}\) and their associated labels \(\mathcal{Y}\) / masks. The memory can be updated with input feature \(X_t\) and estimated output \(Y_t\) at time \(t\);
    \item A target model $\mathrm{C}^t$, which is trained on the memory $\mathcal{M}^t$ at time \(t\), and provides information to decoder $\mathrm{D}$;
    \item Pretrained decoder $\mathrm{D}$ and label encoder $\mathrm{E}$~\cite{bhat2020learning} networks which obtain temporal information from the target model alongside the encoder's output, to generate a fine-grain output mask $Y$ from frame $F$. 
\end{enumerate}
The time index \(t\) is based on input time frame. Thus, at time \(t\), $\mathrm{C}^{t-\Delta_{\mathrm{C}}}$ is updated to $\mathrm{C}^{t}$ on $\mathcal{M}^t$ where $\Delta_{\mathrm{C}}$ is the target model update step. Next, the output $Y_{t+1}$ is estimated from $\mathrm{C}^{t}$, thus $\mathcal{M}^{t}$ can be augmented with pairs ($X_{t+1} , Y_{t+1}$) to create $\mathcal{M}^{t+1}$. 
Potentially, we could update $\mathcal{M}$ at every time frame \(t\), but for practical and computational reasons, we can choose to update the memory every $\Delta_{\mathcal{M}}$ frames, where $\Delta_{\mathcal{M}}$ is the memory update step. An analogous target model update step $\Delta_{\mathrm{C}}$ is considered for updating $\mathrm{C}$. This process is depicted in Figure~\ref{fig1}.

All of the parameters of the VOS model (\(\Xi\)) are first trained offline on a set of training data containing video frames and annotated labels; however, certain parameters of the model need to be updated online at testing time on the extracted features $\mathcal{X}$ of evaluated frames and their associated predicted labels $\mathcal{Y}$ which are kept in the memory $\mathcal{M}$. In particular, let \(\Theta\) be the parameters of target model $\mathrm{C}$, consisting mainly of convolutional filter weights, for $\Theta = \{\theta_l\}_{l=1}^K$ where $K$ is the number of target model parameters. It should be emphasized that $\Theta$ is a rather small subset of the overall parameter set $(\Xi)$, since the target model $\mathrm{C}$ is usually a small convolutional neural network for reasons of efficiency. 
The target model is updated every \(\Delta_{\mathrm{C}}\) frames throughout the video, repeatedly trained on features \(\mathcal{X}\) and associated encoded labels $\mathrm{E}(\mathcal{Y})$ of stored decoder outputs \(\mathcal{Y}\) from  preceding frames. Both \(\mathcal{X}\) and \(\mathcal{Y}\) are stored in memory \(\mathcal{M}\), as shown in Figure~\ref{fig1}.

It is worth noting that $\mathrm{E}$ is a label encoder, generating sub-mask labels from each \(Y\)~\cite{bhat2020learning}. For online training of $\mathrm{C}^{t-\Delta_{\mathrm{C}}}$ at time $t$, every \(Y \in \mathcal{M}^t\) is fed to $\mathrm{E}$ and we seek a trained model $\mathrm{C}^t$ to learn what $\mathrm{E}$ specifies from each \(Y\). That is, the target model acts like a dynamic attention model to generate a set of score maps \(\mathrm{E}\big(\mathrm{C}^t(X)\big)\) in order for the segmentation network ($\mathrm{D}$) to produce the segmented output mask \(Y\) associated with each frame \(F\). The loss function \(L\), which is used for the online training of target model $\mathrm{C}^t$ at time $t$, is
\begin{align}
\label{eq:frtm}
&L(\Theta^t,\mathcal{M}^t) = \\
&\sum_{n=1}^{\abs{\mathcal{M}^t}}\Big\|d_n W_n \Big(\mathrm{E}(Y_n)-\mathrm{E}\big(\mathrm{C}^{t}(X_n)\big)\Big)\Big\|^2_2 
+\sum_{k=1}^{K}\lambda ~{\theta_k^t}^2, \nonumber 
\end{align}
where $\theta^{t}_k \in \Theta^t$ is a parameter of $\mathrm{C}^{t}$ and $\abs{\mathcal{M}^t}$ is the number of feature and mask pairs $\{X,Y\}$ in the memory $\mathcal{M}^{t}$. 

Depending on the overall architecture, $\mathrm{E}$ is an offline / pre-trained label encoder network, as in \cite{bhat2020learning}, or just a pass-through identity function, as in \cite{robinson2020learning}. It is worth noting that the influence and effect of $\mathrm{E}$ is not the focus or interest of this paper. 

In~\cref{eq:frtm},
\(W_n\) is the spatial pixel weight, deduced from \(Y_n\), and $d_n$ is the associated temporal weight decay coefficient. In the loss function $L(\Theta^t,\mathcal{M}^t)$, \(W_n\) balances the importance of the target and the background pixels in each frame, whereas \(d_n\) defines the temporal importance of pair of feature and mask \((X_n,Y_n)\) in memory, typically emphasizing more recent frames~\cite{bhat2020learning}. 

\section{Proposed dataset}
\label{sec:proposed_dataset}

\begin{figure*}[t]
	\begin{center}
		\includegraphics[scale = 0.22]{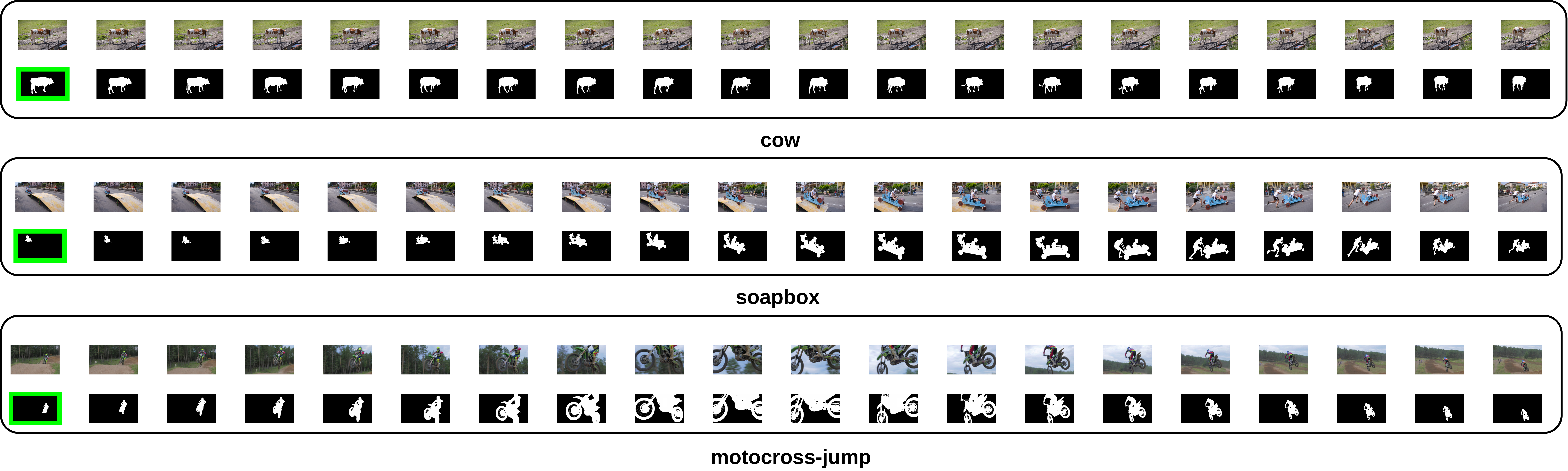}
	\end{center}
	\caption{A set of sub-sampled frames from three videos of the DAVIS16 dataset~\cite{perazzi2016benchmark}, in each case two rows: 
 actual images (top) and segmented objects (bottom). The first video, ``cow'' is the longest in DAVIS16, however there is no significant change between frames. There is a gradual change in appearance in the other two videos. The given annotated (ground-truth) frame in each video is highlighted in green.}
	\label{fig2}
\end{figure*}
As shown in Figure~\ref{fig1},  online VOS assumes the change in each video sequence to be gradual, meaning that a constant size of memory \(\mathcal{M}^t\) has an adequate capacity to update the target model \(\mathrm{C}^{t-\Delta_{\mathrm{C}}}\) to \(\mathrm{C}^{t}\) for segmenting the current frame $F_{t+1}$. In the ideal case, where the samples in a video sequence are independent and identically distributed (i.i.d.), machine learning problems are made significantly easier, since there is then no need to handle distributional drift and temporal dependency in VOS. However, i.i.d. assumption is not valid in video data.

Figure~\ref{fig2} shows three video sequences from the DAVIS2016 dataset, where we can see that target objects do not have an abrupt change through video frames. Objects could have small changes, such as in the ``cow'' video (the longest video  in DAVIS2016 at $104$ frames), and the other two videos (soapbox and motocross-jump) possess variations in object appearance, however the changes are gradual.  As a result, for such datasets the identically distributed assumption of frames is usually valid, particularly for short videos.  It is thus worth mentioning that the YouTube-VOS18 sequences are even shorter than those in DAVIS16 and DAVIS17, where the longest video in the validation set of YouTube-VOS18 has $36$ frames. 

The semi-supervised VOS approaches maintain the i.i.d.\ assumption for video sequences, despite the fact that this assumption is clear not valid in all video sequences, particularly longer ones. It is precisely for this  reason that state-of-the-art semi-supervised VOS models are not expected to have a similar performance on long video datasets~\cite{cheng2022xmem}.

\begin{figure*}[t]
	\begin{center}
		\includegraphics[scale = 0.22]{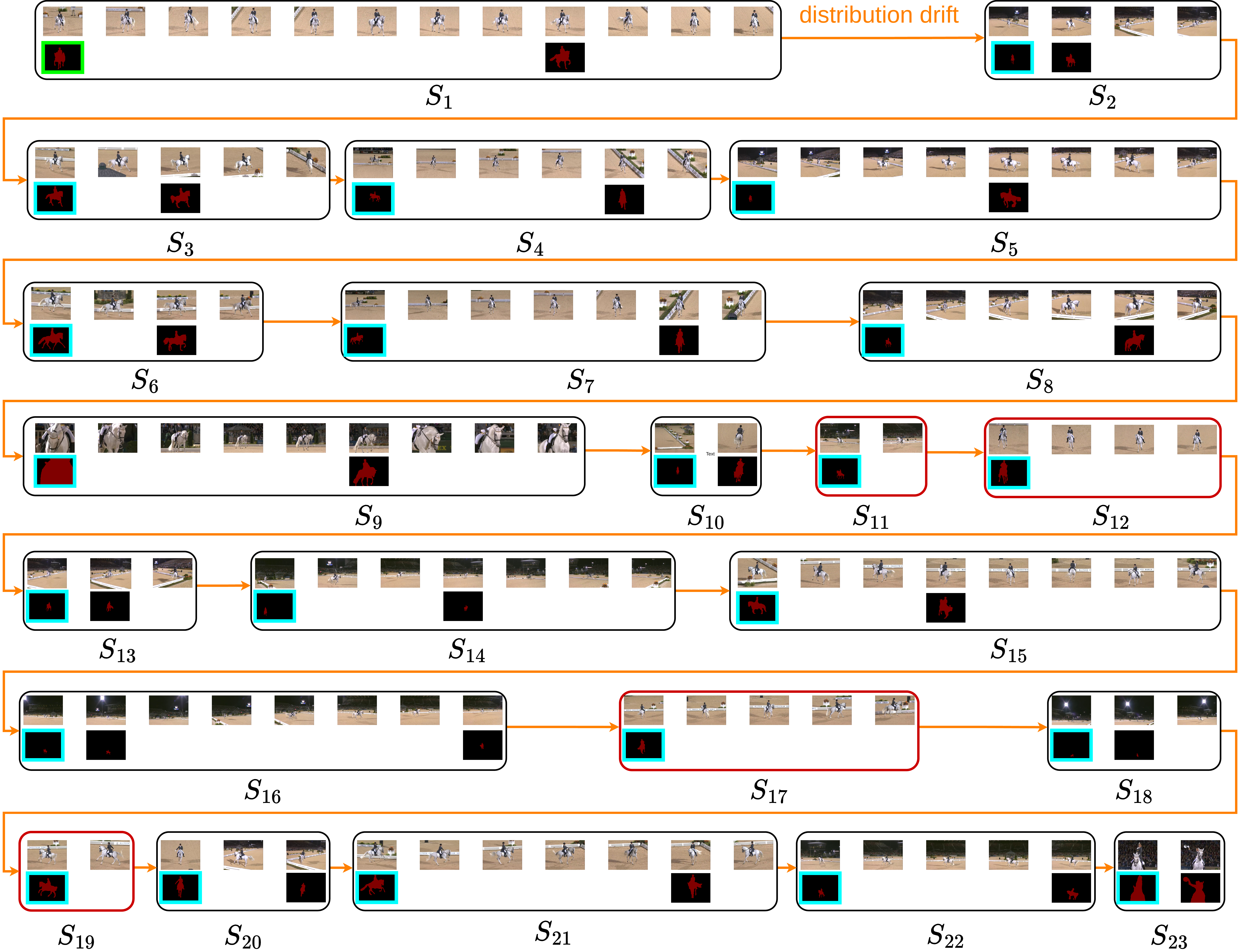}
	\end{center}
	\caption{A subset of frames from ``dressage'' video of the Long Videos dataset~\cite{liang2020video}. The video consists of $23$ sub-chunks that are separated from each other by significant distributional drifts or discontinuities. The lower (sparse) row, in each set, show the annotated frames. The  annotations provided by~\cite{liang2020video} are shown without a border, whereas the annotated masks added via this paper, and made available via the CLVOS23 dataset, are shown with blue borders. The four sub-chunks that are missing from the Long Videos dataset are encircled in red.}
	\label{fig3}
\end{figure*}

Figure~\ref{fig3} shows the  ``dressage'' video from the Long Videos dataset~\cite{liang2020video}, the dataset consisting of three long sequences with a total of $7411$ frames.  As is clear from Figure~\ref{fig3}, an i.i.d.\ assumption is not at all valid on ``dressage'' video, because of the $22$ substantial distribution drifts which take place, a behaviour which is much more closely aligned with the {\em non}-i.i.d.\ assumption of continual learning. However, this new continual learning-based interpretation of the long video sequences is discussed for the first time in VOS and continual learning. As the evaluation label mask is chosen uniformly in the Long Videos dataset, it does not show how well a VOS solution handles sudden shifts in the target's appearance.
Alternatively, we propose annotating the frames for the evaluation based on the distribution drift that occurs in each video sequence. 

Figure~\ref{fig3} shows $23$ sub-chunks of videos in the ``dressage'' video of the Long Videos dataset. Each sub-chunk is separated from its previous and next sub-chunks based on the distribution drifts. When an online or offline event, such as a sports competition, is recorded using multiple cameras, these distribution drifts are common in media-provided videos.
As a result, in our proposed dataset, we first utilize the following strategy to select candidate frames for annotation and evaluation. 
\begin{itemize}
    \item We select the first frame of each sub-chunk \(S\). It is interesting to see how VOS models handle the distribution drift that happens in the sequence, which is arriving a new task in continual learning.
    \item The last frame of each sequence is also selected. The first frame ground truth label mask is given to the model as it is set in the semi-supervised VOS scenario.
    \item One frame from the middle of each sub-chunk is also selected for being annotated.
\end{itemize}
As shown in Figure~\ref{fig3}, selecting the annotated frames uniformly will cause some small sub-chunks ($S_{11},S_{12},S_{17},S_{19}$) to be missed in the evaluation.
For CLVOS23, in addition to the $3$ videos from the Long Videos dataset, we added the other $6$ videos described in Table~\ref{tab1}. All frames of the $6$ new added videos are extracted with the rate of $15$ Frames Per Second (FPS). To ensure that all distribution drifts are captured, we only annotate the first frame of each sub-chunk in the Long Videos dataset and add them to the uniformly selected annotated frames.
The proposed dataset has following advantages over the Long Videos dataset~\cite{liang2020video}.
\begin{itemize}
    \item It added $5951$ frames to $7411$ frames of the Long Videos dataset.
    \item CLVOS23 increased the number of annotation frames from $63$ in the Long Videos dataset to $284$.
    \item It increases the number of videos from $3$ to $9$.
    \item The selected annotated frames are chosen based on the distribution drift that happens in the videos (sub-chunks) rather than being uniformly selected.
\end{itemize}
It is worth noting that for a long VOS dataset, it is very expensive and sometimes unnecessary to annotate all the frames of videos for evaluation. It is worth mentioning that We utilized the Toronto Annotation Suite~\cite{torontoannotsuite} to annotate the selected frames for evaluation.  The frames of new $6$ videos were resized to have a height of $480$ pixels. The width of each frame is defined as proportionate to its height. The link to access to the dataset is provided.\footnote{\url{https://github.com/Amir4g/CLVOS23}} 

\begin{table}
  \centering
  \resizebox{\linewidth}{!}{
  \begin{tabular}{l|c|c|c}
    \toprule
    \centering
    Video name &\#Sub-chunks (tasks) &\#Frames & \#Annotated frames\\
    \midrule
    dressage &$23$&$3589$&$43$ \\
    blueboy &$27$&$1416$&$47$\\
    rat &$22$&$2606$&$42$ \\
    car &$18$&$1109$&$37$ \\
    dog &$12$&$891$&$25$ \\
    parkour &$24$&$1578$&$49$ \\
    skating &$5$&$778$&$11$\\
    skiing &$5$&$692$&$11$ \\
    skiing-long &$9$&$903$&$19$ \\
    \bottomrule
  \end{tabular}}
  \caption{Each video sequence's specifications in the proposed CLVOS23 dataset. The first three videos (Dressage, Blueboy, and Rat) are taken directly from the Long Videos dataset~\cite{liang2020video} and we added additional annotated ground-truth frames to each of them to make them more appropriate for continual learning.} 
  \label{tab1}
\end{table}

\section{Proposed method}
\label{sec:proposed_method}

\begin{figure*}[t]
	\begin{center}
		\includegraphics[scale = 0.21]{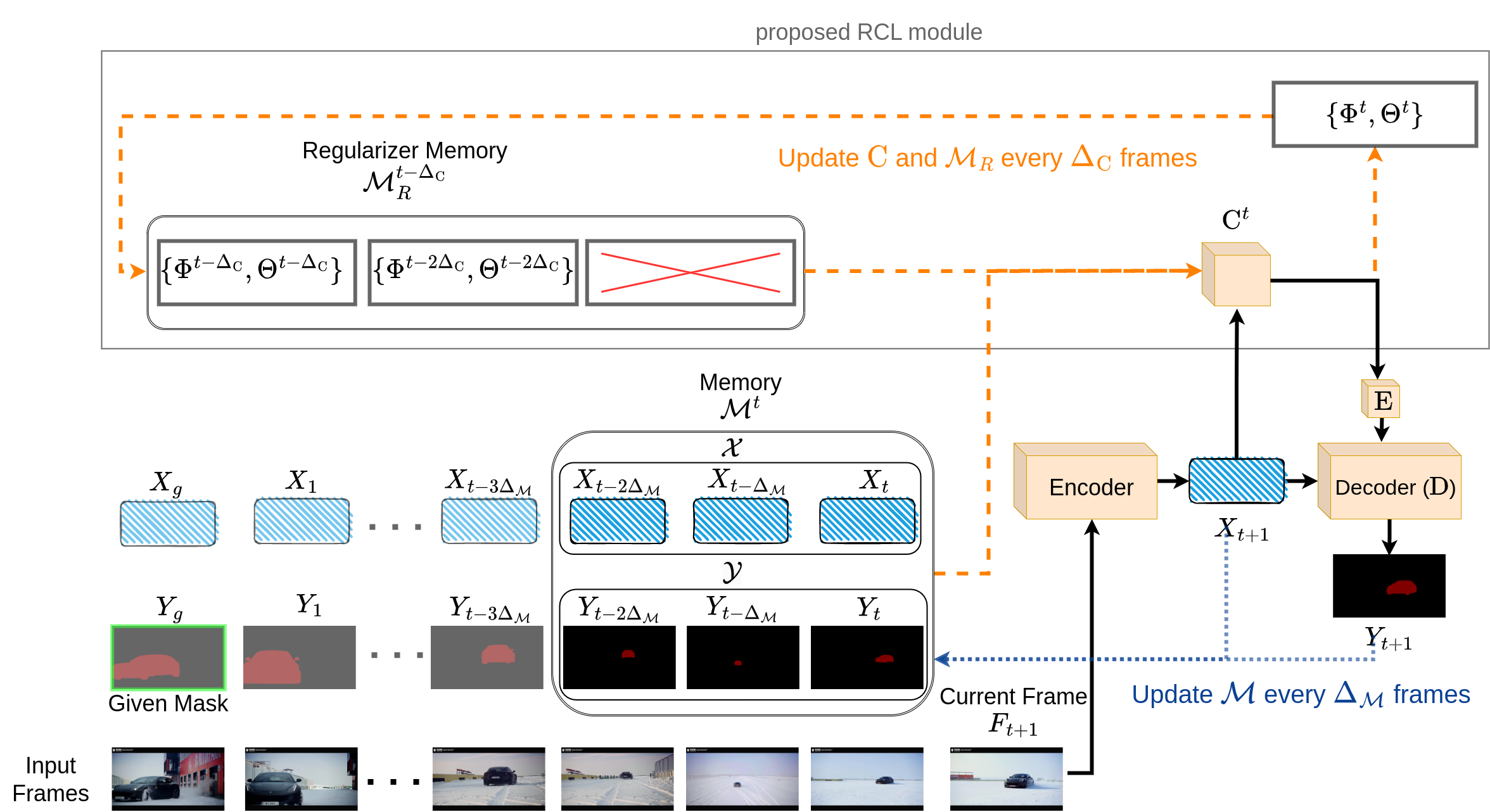}
	\end{center}
	\caption{The proposed online VOS framework, with the proposed RCL approach: At time \(t\), the process of updating $\mathrm{C}^{t-\Delta_{\mathrm{C}}}$ on $\mathcal{M}^t$ is regularized by all pairs of the target model's parameters and their associated importance \(\{\Phi,\Theta\}\) in the regularizer memory $\mathcal{M}_R^{t-\Delta_{\mathrm{C}}}$ as shown in \cref{eq2}. After updating $\mathrm{C}^{t-\Delta_{\mathrm{C}}}$ to $\mathrm{C}^{t}$, $\mathcal{M}_R^{t-\Delta_{\mathrm{C}}}$ is updated using \(\{\Phi^t,\Theta^t\}\) calculated from $\mathrm{C}^{t}$.}
	\label{fig-RCL}
\end{figure*}
A continual learning system should have a limited constant memory which is essential for a bounded system working on an infinite sequence of data. Thus, we focus on addressing continual learning using the memory-based VOS models and among them we are interested in the online VOS approaches, where part of the model ($\mathrm{C}$) is updating on a constant size memory $\mathcal{M}$. 

The LWL method~\cite{bhat2020learning}, which is an extension over the well-known FRTM framework~\cite{robinson2020learning} benefits from a label encoder network $\mathrm{E}$ that tells the target model $\mathrm{C}$ what to learn~\cite{bhat2020learning}. In this article, LWL has been chosen as the online VOS baseline method.
The framework structure that is explained in Figure~\ref{fig1} is followed by LWL, where encoder, decoder $\mathrm{D}$, and the label encoder $\mathrm{E}$ are all trained offline; consequently, we do not make any modifications to these components by implementing the proposed solution.

The proposed regularization-based continual learning (RCL) method is inspired by the EWC~\cite{kirkpatrick2017overcoming} algorithm,  where the network parameters \(\Theta\) of the target model \(\mathrm{C}\) in  LWL are regularized to preserve the important parameters and prevent modification during the target model updating steps. The importance of each parameter \(\theta_k\) is associated with the magnitude of its related gradient \(\phi_k\) during the preceding update steps. Therefore, during each updating (online learning) step \(t\), the training parameters $\Theta^t$ are regularized by the magnitude of the gradients of the target models' parameters $\Phi = \{\phi_k\}_{k=1}^{K}$ and the updated model's parameters \(\Theta= \{\theta_k\}_{k=1}^{K}\) of preceding updates, which are stored in the regularizer memory $\mathcal{M}_{R}$. 

Thus, for all features \(\mathcal{X}\) and their related output masks \(\mathcal{Y}\) in the memory \(\mathcal{M}^t\), the target model \(\mathrm{C}^t\) with  parameters \(\Theta^t\), and the regularizer memory $\mathcal{M}_R^{t-\Delta_{\mathrm{C}}}$, the following loss function defined in~\cref{eq2} is used for training the target model of LWL:
\begin{align}
\label{eq2}
&L_{R}(\Theta^t,\mathcal{M}^t,\mathcal{M}_R^{t-\Delta_{\mathrm{C}}}) =
\\
&L(\Theta^t,\mathcal{M}^t) +
\lambda \sum_{j=1}^{\abs{\mathcal{M}_R^{t-\Delta_{\mathrm{C}}}}}\Phi^j\Big{|}\Big{|}\Theta^t-\Theta^{j}\Big{|}\Big{|}^2
\nonumber
\end{align}
where the loss function \(L\) is described in~\cref{eq:frtm}, \(\lambda\) controls the regularisation term, and $\abs{\mathcal{M}_R^{t-\Delta_{\mathrm{C}}}}$ shows how many pairs of $\{\Theta, \Phi\}$ have been stored in $\mathcal{M}_{R}$ so far.
The loss function in~\cref{eq2} is used to update the target model, and it regularizes the target model training to preserve its previously learned knowledge. The proposed RCL method is depicted in Figure~\ref{fig-RCL}. As illustrated in this figure, the proposed RCL can be added to any online VOS method and improve its performance as shown in Section~\ref{sec:experimental result}.

It is worth noting that the memory \(\mathcal{M}\) is initialized by the encoded features of the given frame \(F_g\) and its provided ground-truth mask \(Y_g\) as defined in a semi-supervised VOS scenario. 

One drawback of the proposed regularization-based method is that it needs to store the parameter importance \(\Phi^t\) and the parameters of the target model \(\Theta^t\) after each online updating step \(t\); however, a limited number of stored pairs of \(\{\Phi,\Theta\}\) are enough to regularize the updating step of the target model \(\mathrm{C}^t\). 

Additionally, for a small target model \(\mathrm{C}\), it is feasible to calculate and store the \(\Phi\) and \(\Theta\) during the updating step; however, it is a real challenge for a larger target model.
 
\section{Experimental Result}
\label{sec:experimental result}

A fixed setup is used for the evaluated methods, with maximum memory sizes of \(N=32\) for LWL and LWL-RCL as suggested in LWL's original publication. For all experiments, the target model $\mathrm{C}$ is updated for three epochs on the memory $\mathcal{M}$ in each updating step to have a fair comparison with the baseline. The target model is updated every time the memory is updated, following the proposed setup in~\cite{cheng2022xmem}.

The memory \(\mathcal{M}^0\) is initialized by the given ground truth frame \(F_g\). In all of the experiments, as suggested in the semi-supervised online VOS baseline (LWL), the information extracted from \(F_g\) is preserved and is used throughout the evaluation of other frames in the video sequence. In the proposed method, the same concept is followed where in the proposed regularisation-based LWL, the importance parameters \(\Phi^0\) and the parameters  \(\Theta^0\) related to the training of the target model \(\mathrm{C}\) on $X_g$ and $Y_g$ are kept in \(\mathcal{M}_R\).

In the RCL method, $\lambda$ is set to $5$ and the maximum size of $\mathcal{M}_R$ is set to $20$. We validate these hyper-parameter using cross validation. In LWL, the target model $\mathrm{C}$ is a small one layer convolutional neural network. Additionally, the same pretrained decoder $\mathrm{D}$ and encoder models are used for all experiments of LWL. To measure the effectiveness of the proposed method, consistent with the standard DAVIS protocol~\cite{perazzi2016benchmark} the mean Jaccard \(\mathcal{J}\) index, mean boundary \(\mathcal{F}\) scores, and the average of \(\mathcal{J}\& \mathcal{F}\) are reported for all evaluated methods. The speed of each method is reported on the DAVIS16 dataset~\cite{robinson2020learning} in units of Frames Per Second (FPS).
All experiments were performed using one NVIDIA V100 GPU.

The effectiveness of the proposed regularization-based continual learning method (RCL) is evaluated by augmenting an online VOS framework (LWL); however, the proposed method can be extended to any online VOS method having a periodically-updated network model, as in Figure~\ref{fig1}. 

Table~\ref{tbl2} shows the results of the selected baseline (LWL) and the augmented baseline with the proposed regularization-based method (RCL) on the Long Video dataset~\cite{liang2020video}, and the proposed CLVOS23 dataset.
Here, six experiments with six different memory and target model update step sizes \(\Delta_{\mathrm{C}} \in \{1,2,4,6,8,10\}\) are conducted ($\Delta_{\mathcal{M}}=\Delta_{\mathrm{C}}$), where, the memory $\mathcal{M}^t$ is updated after each target model \(\mathrm{C}^{t-\Delta_{\mathrm{C}}}\) update to \(\mathrm{C}^{t}\). For reference,  the means and standard deviations of six runs of two  competing methods (LWL and LWL-RCL) are reported in Table~\ref{tbl2}. As it is represented in Table~\ref{tbl2}, CLVOS23 is a more difficult VOS dataset in comparison to the Long Videos dataset, since LWL has lower performance on CLVOS23. Additionally, the proposed RCL improves LWL on CLVOS23 more than the Long Videos dataset, which shows CLVOS23 is a more appropriate dataset for evaluating online, continual learning-based contributions.

Furthermore, looking at the standard deviations reported in Table~\ref{tbl2}, the proposed regularization-based method decreases the standard deviation of reported results with different memory and target model step sizes \(\Delta_{\mathrm{C}} \in \{1,2,4,6,8,10\}\). This indicates that the proposed method is more robust against selecting different frame rates for updating the target model $\mathrm{C}$.

Table~\ref{tbl3} shows the results of the selected baseline on two short VOS datasets (DAVIS16 and DAVIS17). The results show that the proposed RCL method does not have any negative effects on the accuracy of the baseline method (LWL); however, it affects the speed of the baseline since it needs to recalculate the regularization term in \cref{eq2} in every epoch of the updating step.

It is worth mentioning that we use the suggested hyper-parameters in the original paper of LWL~\cite{bhat2020learning}; nevertheless, the used hyper-parameters are not necessarily the best parameters for LWL on long video datasets, and it is possible to improve the performance of the baseline method on the evaluated dataset by only making some small changes to LWL. The objective of this article is to provide a continual learning-based VOS dataset and a method that improves any online VOS approaches that struggle with forgetting on long video sequences with abrupt changes in the target object's appearance.
\begin{table}
\centering
\resizebox{\linewidth}{!}{
\begin{tabular}{l|ccc|ccc}
\toprule
    \multirow{2}{*}{Method}  &  \multicolumn{3}{c|}{Long Videos~\cite{liang2020video}} & \multicolumn{3}{c}{CLVOS23}
    \\
    \cmidrule{2-7}
                             &\(\mathcal{J}\) & \(\mathcal{F}\) & \(\mathcal{J}\&\mathcal{F}\) & \(\mathcal{J}\) & \(\mathcal{F}\) & \(\mathcal{J}\&\mathcal{F}\)\\

\midrule
			LWL~\cite{bhat2020learning} &   78.0\(\pm\)4.3 & 81.6\(\pm\)4.2 & 79.8\(\pm\)4.2 & 68.1\(\pm\)2.2 & 71.9\(\pm\)2.4 & 70.0\(\pm\)2.3\\
   			LWL-RCL (ours)&                 79.8\(\pm\)3.0 & 82.7\(\pm\)3.2 & 81.3\(\pm\)3.1 & 70.4 \(\pm\)1.9 & 74.33\(\pm\)2.0 & 72.4\(\pm\)2.0\\
\bottomrule
\end{tabular}}
\caption{Performance analysis of the evaluated methods against the validation set of the Long Videos and proposed CLVOS23 datasets.}
\label{tbl2}
\end{table}

\begin{table}
\centering
\resizebox{\linewidth}{!}{
\begin{tabular}{l|ccc|ccc|c}
\toprule
    \multirow{2}{*}{Method}  &  \multicolumn{3}{c|}{DAVIS17} & \multicolumn{3}{c|}{DAVIS16}& \multirow{2}{*}{FPS}
    \\ 
    \cmidrule{2-7}
  &\(\mathcal{J}\) & \(\mathcal{F}\) & \(\mathcal{J}\&\mathcal{F}\) &
  
  \(\mathcal{J}\) & \(\mathcal{F}\) & \(\mathcal{J}\&\mathcal{F}\)& \\ 
\midrule
			LWL~\cite{bhat2020learning} &  77.1 & 82.9 &80.0& 87.3 & 88.5 &87.9& 18.15
			\\
   			LWL-RCL (ours)&  77.1 & 82.9 & 80.0 & 87.3 & 88.5 &87.9& 14.47 \\
\bottomrule
\end{tabular}}
\caption{Performance analysis of the evaluated methods against validation sets of the DAVIS16 and DAVIS17 datasets.}
\label{tbl3}
\end{table}

\section{Conclusion}
\label{sec:conclusion}
In this article, we presented a dataset called CLVOS23 to examine the capability of semi-supervised VOS approaches to deal with the forgetting of past frames' learning, and we frame this problem as a continual learning challenge.
To help online VOS methods get around memory limitations without sacrificing accuracy, we also proposed adding a regularization-based module to them. The proposed modules can be added to any existing online VOS framework that is already in place to make it more efficient and resistant to distribution drifts that can happen during long video clips, while keeping or even improving performance accuracy. The changes we made to the standard procedure for online VOS made it more accurate on long videos, according to our results. Furthermore, on the short video datasets (DAVIS16, DAVIS17) where the object's appearance does not suddenly change, the proposed methods do not outperform the baselines.

\section*{Acknowledgments}
We appreciate the generous support provided by Microsoft Office Media Group and NSERC Alliance for this research project.




{\small
\bibliographystyle{ieee_fullname}
\bibliography{camera_ready}
}

\end{document}